\documentclass{article}
\usepackage{spconf,amsmath,graphicx}
\usepackage{algorithm}
\usepackage{algorithmic}
\usepackage{amsthm}
\usepackage{amsfonts}
\usepackage{amsmath}

\usepackage{subcaption}
\usepackage{booktabs}
\usepackage{xspace}
\usepackage{multirow}
\usepackage{textcomp}
\usepackage{siunitx}
\usepackage{xcolor}
\usepackage{enumitem}
\usepackage[normalem]{ulem}
\usepackage{stfloats}

\newcommand{\our}{\mbox{STAug}\xspace}
\newcommand{\timeabl}{\mbox{STAug-noTime}\xspace}
\newcommand{\freqabl}{\mbox{STAug-noFreq}\xspace}

\title{Towards Diverse and Coherent Augmentation \\ for Time-Series Forecasting}

\name{Xiyuan Zhang\textsuperscript{\rm 1}, Ranak Roy Chowdhury\textsuperscript{\rm 1}, Jingbo Shang$^{*}$\textsuperscript{\rm 1}, Rajesh Gupta$^{*}$\textsuperscript{\rm 1}, Dezhi Hong$^{*}$\textsuperscript{\rm 2} \thanks{$^*$Corresponding authors.}}

\address{\textsuperscript{\rm 1} University of California,  San Diego, La Jolla, CA, USA \\
\textsuperscript{\rm 2} Amazon$^{\dagger}$ \thanks{$^*$Work unrelated to Amazon.}}

\begin{document}
%
\maketitle

\begin{abstract}
Time-series data augmentation mitigates the issue of insufficient training data for deep learning models.
Yet, existing augmentation methods are mainly designed for classification, where class labels can be preserved even if augmentation alters the temporal dynamics. We note that augmentation designed for forecasting requires \emph{diversity} as well as \emph{coherence} with the original temporal dynamics. 
As time-series data generated by real-life physical processes exhibit characteristics in both the time and frequency domains, we propose to combine Spectral and Time Augmentation (STAug) for generating more diverse and coherent samples. 
Specifically, in the frequency domain, we use the Empirical Mode Decomposition to decompose a time series and reassemble the subcomponents with random weights.
This way, we generate diverse samples while being coherent with the original temporal relationships as they contain the same set of base components.
In the time domain, we adapt a mix-up strategy that generates diverse as well as linearly in-between coherent samples.
Experiments on five real-world time-series datasets demonstrate that \our outperforms the base models without data augmentation as well as state-of-the-art augmentation methods.
\end{abstract}

\section{Introduction}

Deep learning has been successful in various time-series applications given enormous amount of data to train. However, time-series data collected through real-world sensors is often marked by irregular samples with missing values due to collection difficulties. Such data {\em scarcity} commonly observed in time-series data can significantly degrade the performance of deep learning methods that would otherwise perform well.

\begin{figure}[t]
      \centering
      \begin{subfigure}[b]{0.155\textwidth}
          \centering
          \includegraphics[width=\textwidth]{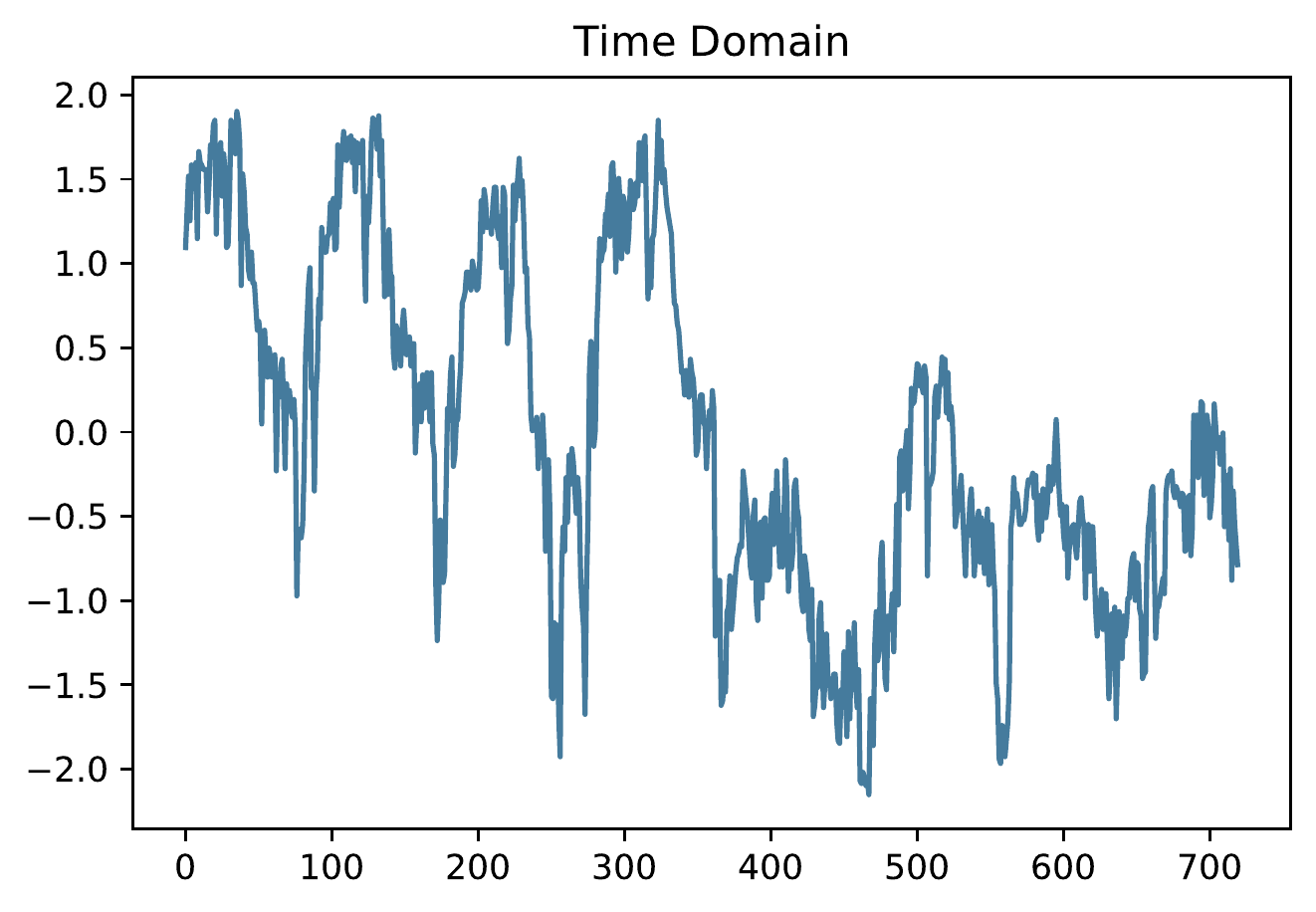}
          \caption{Original}
          \label{fig:ex-ori}
      \end{subfigure}
      \begin{subfigure}[b]{0.155\textwidth}
          \centering
          \includegraphics[width=\textwidth]{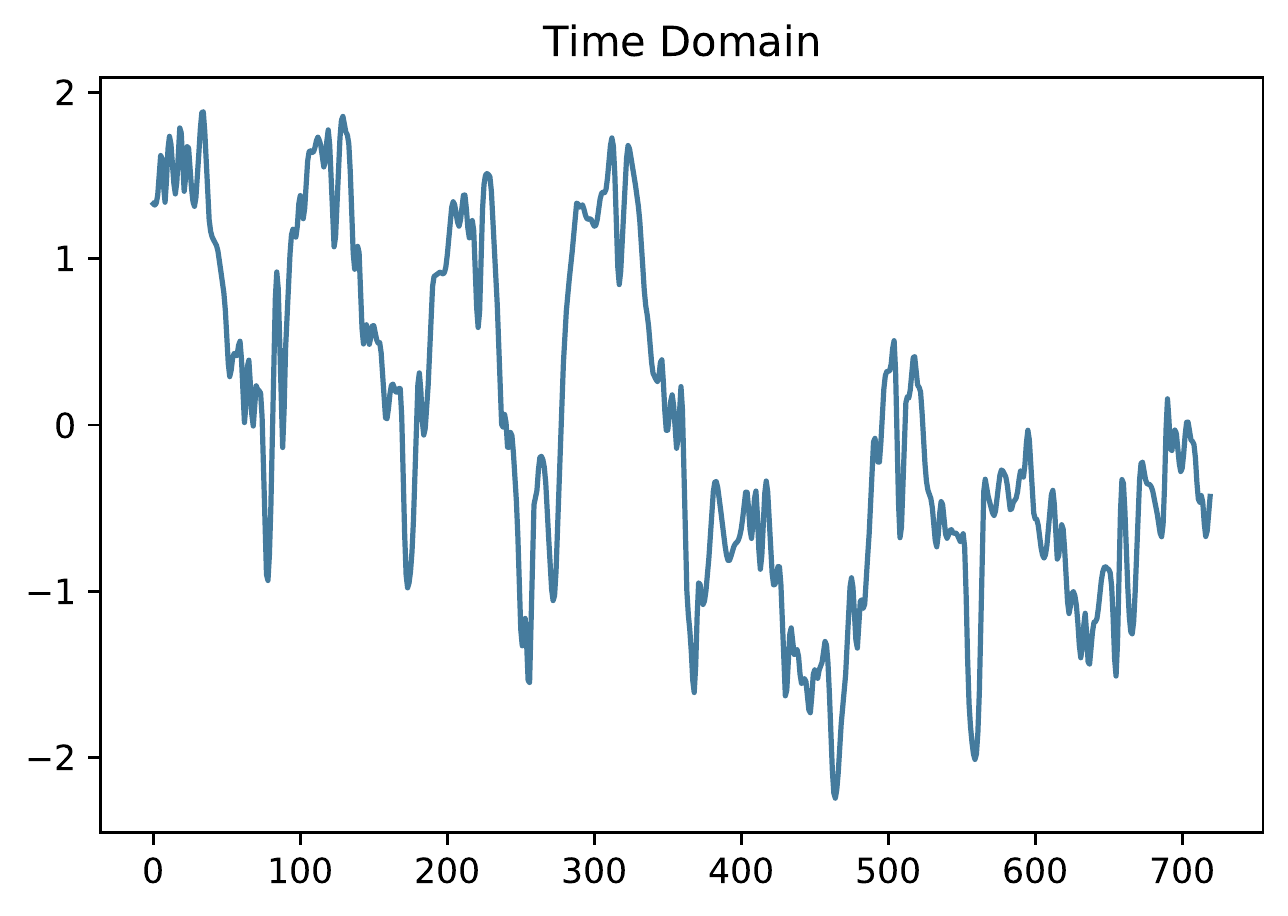}
          \caption{Filtering}
          \label{fig:ex-ww}
      \end{subfigure}
      \begin{subfigure}[b]{0.155\textwidth}
          \centering
          \includegraphics[width=\textwidth]{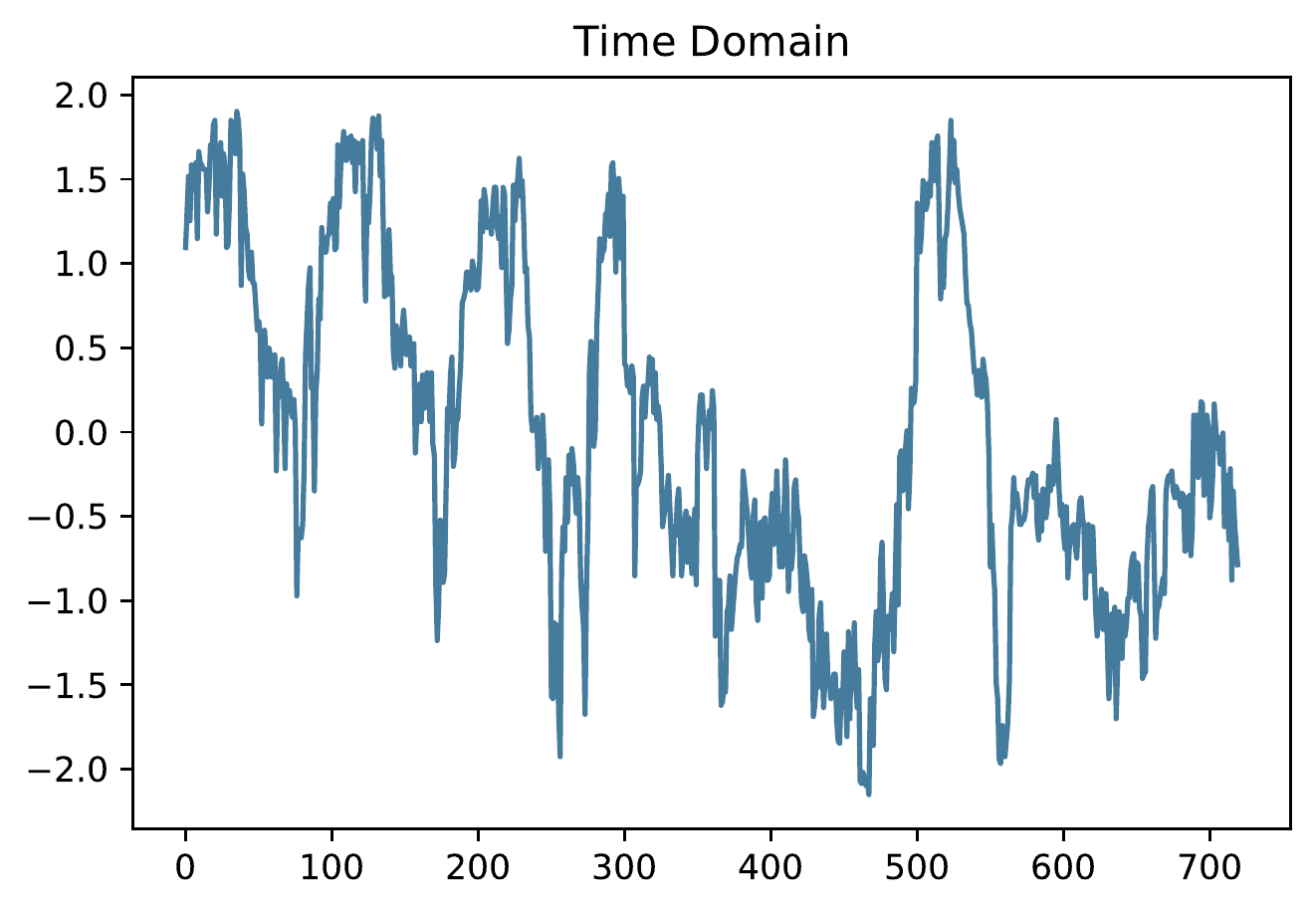}
          \caption{Permutation}
          \label{fig:ex-perm}
      \end{subfigure}
         \caption{Visualization of original and augmented time series from ETTm2. Augmentation methods often (b) miss diversity or (c) miss coherence with the original temporal dynamics.} 
         \label{fig:ex}
\end{figure}

A rich line of research tries to address this problem through data augmentation, that is to generate synthetic data points to augment the original dataset~\cite{um2017data,le2016data,bergmeir2016bagging,forestier2017generating,esteban2017real,che2017boosting,steven2018feature,lim2018doping,yoon2019time,nam2020data,hu2020datsing,kang2020gratis}. 
However, existing augmentation methods are mainly designed for classification, where augmented samples remain effective as long as they preserve the class labels. 
We note that augmentation designed for forecasting requires both \emph{diversity} and \emph{coherence} with the original temporal dynamics. 
Yet, existing augmentation methods generate samples that often miss one of the criteria (Figure~\ref{fig:ex}). For example, filtering-based methods are deterministic processes that produce a fixed set of synthetic samples by removing noises. Permutation-based methods change the temporal order of the original series, worsening forecasting performance.

Moreover, time-series data generated by real-life physical processes exhibit characteristics both in the time and frequency domains that are not available in other data modalities like image and text.
Therefore, temporal dynamics can be best captured through a joint consideration of time domain that carries changes over time and frequency domain that conveys periodic patterns. 
By contrast, existing augmentation methods mostly generate data in one domain, ignoring the complementary strengths of both domains.

\begin{figure}[t]
\centering
\includegraphics[width=\linewidth]{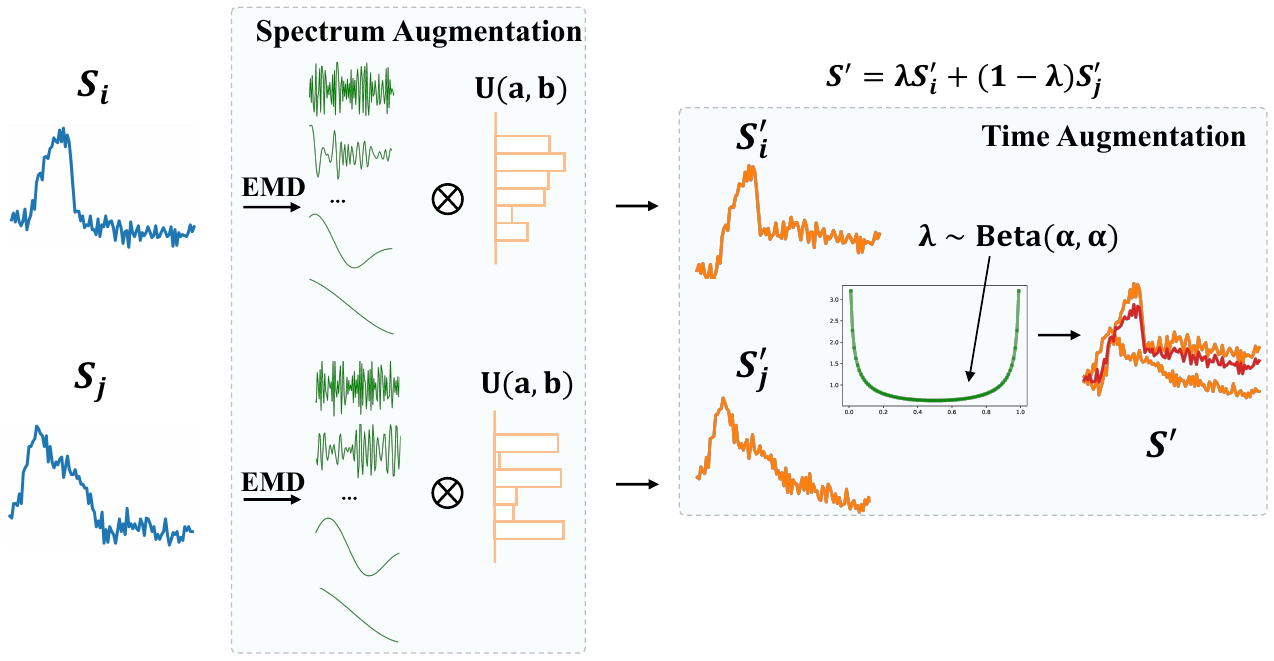}
\vspace{-4mm}
\caption{Overview of \our. For frequency-domain augmentation, we decompose two random series $\mathbf{S_i}, \mathbf{S_j}$ (marked in blue) with Empirical Mode Decomposition (EMD), and then reassemble the subcomponents with random weights sampled from uniform distribution $U(a,b)$ to obtain $\mathbf{S'_i}, \mathbf{S'_j}$ (marked in orange). In the time domain, we further linearly mix $\mathbf{S'_i}, \mathbf{S'_j}$ to obtain the augmented series $\mathbf{S'}$ (marked in red) for training. 
}
\label{fig:main}
\end{figure}

We propose \emph{\our} (Figure~\ref{fig:main}), by combining Spectral and Time Augmentation for time-series forecasting task.
In the frequency domain, we first apply the Empirical Mode Decomposition (EMD)~\cite{huang1998empirical}
to decompose time series into multiple subcomponents, each representing a certain pattern embedded in the data. 
We then reassemble these subcomponents with random weights to generate new synthetic series.
This offers a \emph{principled} way of augmentation as it generates diverse samples while maintaining the same basic set of subcomponents.
We adopt EMD for the frequency information as it better captures patterns for non-stationary time series compared with Fourier transform.
In the time domain, we adapt a mix-up strategy~\cite{zhang2017mixup} to learn linearly in-between randomly sampled pairs of training series, which produces varied and coherent samples. We evaluate \our on five real-world time-series datasets, and the method demonstrates state-of-the-art performance compared with existing augmentation methods.

\section{Related Work}
Existing time-series data augmentation techniques mainly fall into four categories~\cite{wen2020time,iwana2021empirical} as follows. 

\emph{Basic random operations} include time-domain transformation, frequency-domain transformation and time-frequency domain transformation. 
Time-domain transformation contains scaling, rotation, jittering~\cite{um2017data}, slicing, warping~\cite{le2016data,ma2021joint}, etc. 
For frequency-domain transformation, RobustTAD~\cite{gao2020robusttad} in the frequency domain makes perturbations in both magnitude and phase spectra.
For time-frequency domain transformation, time-frequency features are generated from Short Fourier Transform (STFT), and then local averaging together with feature vector shuffling are applied for augmentation~\cite{steven2018feature}.
SpecAugment~\cite{park2019specaugment} proposes augmentation in Mel-Frequency by combining warping, masking frequency channels and masking time step blocks together.

\emph{Decomposition-based augmentation} leverages the Seasonal-Trend Decomposition (STL) ~\cite{cleveland1990stl} or Empirical Mode Decomposition (EMD)~\cite{huang1998empirical} to extract patterns for generating synthetic samples. 
Bagging Exponential Smoothing method~\cite{bergmeir2016bagging,kegel2018feature,bandara2021improving} uses Box–Cox transformation followed by STL decomposition, and bootstraps the reminder to assemble new series.
STL decomposition components can also be adjusted and combined with a stochastic component generated by statistical models~\cite{kegel2018feature,bandara2021improving}.
Nam et al.~\cite{nam2020data} decomposes series using EMD into components from high frequency to low frequency, and adds the residue each time one IMF occurs. This can be viewed as a special case of \our with weights equal to one for low-frequency components and zero for high-frequency components, essentially only filtering out high-frequency noise. Moreover, the method does not benefit from time-domain augmentation.

For \emph{pattern mixing method}, DBA calculates weighted average of multiple time series under DTW as new samples~\cite{forestier2017generating,bandara2021improving}. Mix-up~\cite{zhang2017mixup} constructs new examples through linear interpolation of both features and labels, but such interpolation methods mainly focus on classification tasks. 

\emph{Generative method} models underlying distribution of the dataset for generation, including both statistical generative models~\cite{cao2014parsimonious,kang2020gratis} and deep generative adversarial network (GAN)~\cite{goodfellow2014generative} based models~\cite{esteban2017real,che2017boosting,ramponi2018t,lim2018doping,yoon2019time,hu2020datsing}. 

In this work, we combine decomposition-based augmentation method in the frequency domain and pattern mixing augmentation method in the time domain to find diverse samples that preserve the original data characteristics.
\section{Methodology}

\subsection{Overview}
We focus on multivariate time series forecasting task. A multivariate time series sequence of length $T$ and feature number $c$ is denoted as $\mathbf{S}=[\mathbf{s_1},\dots,\mathbf{s_t},\dots,\mathbf{s_T}] \in \mathbb{R}^{c \times T}$, where $\mathbf{s_t} = [s_1, \cdots, s_c]^T \in \mathbb{R}^c$. In a forecasting task, we only observe history values $\mathbf{H}$ up to timestamp $d<T$: $\mathbf{H}=[\mathbf{s_1},\dots,\mathbf{s_d}] \in \mathbb{R}^{c \times d}$, and the goal is to forecast future values $\mathbf{F}$ at timestamp $d+1,\dots,T$: $\mathbf{F}=[\mathbf{s_{d+1}},\dots,\mathbf{s_T}] \in \mathbb{R}^{c \times (T-d)}$, where $\mathbf{S} = [\mathbf{H}, \mathbf{F}]$. During training, we have full access to both $\mathbf{H}$ and $\mathbf{F}$, and the training objective is to  learn a model $g$ that forecasts $\mathbf{F}$ given $\mathbf{H}$ for each $\mathbf{S}$ in the training set. During testing, we have access to \emph{only} $\mathbf{H}$ and input $\mathbf{H}$ to model $g$ to obtain predictions for the future part.

Our method \our comprises augmentation in both the frequency domain and time domain. In the frequency domain, we first apply empirical mode decomposition to obtain a set of components. Then during each iteration, these components are re-combined with random weights to construct a new synthetic series. Then, we adapt mix-up as a time-domain augmentation. We linearly interpolate two randomly re-combined series to obtain the final augmented series. The augmented series are fed into the forecasting model for updating gradient. The EMD components of different series could be pre-computed, and \our only requires randomly re-combining components or series during training, which introduces minimal computational overhead. 

\subsection{Frequency-Domain Augmentation}
The Empirical Mode Decomposition (EMD) \cite{huang1998empirical} method was originally designed to analyze nonlinear and non-stationary data, whose constituent components are not necessarily cyclic. EMD preserves temporal information in contrast with Fourier transform, and is data-driven compared with the linear wavelet transform. It decomposes a signal into a finite number of Intrinsic Mode Functions (IMF).
The first several IMFs usually carry components of higher frequency (e.g., noise), while the last several IMFs represent the low-frequency trend information embedded in the sequence. Therefore, EMD provides a principled way to decompose a signal into multiple components, and each of these components represents certain patterns embedded in the original signal.

After EMD, the original sequence could be written as 

\begin{equation}
\mathbf{S} = \sum_{i=1}^n \mathrm{IMF}_i + \mathbf{R}. 
\end{equation}

With a list of $n$ decomposed IMFs $\{\mathrm{IMF}_1,\dots,\mathrm{IMF}_n\}$ and residual $\mathbf{R}$, we apply a random vector $\mathbf{w}=[w_1,\dots,w_n]^T$ as weights to re-combine these IMFs as $\mathbf{S'}$:

\begin{equation}
    \mathbf{S'} = \sum_{i=1}^n w_i \cdot \mathrm{IMF}_i
\end{equation}

where $w_i$ is sampled from uniform distribution $U(0,2)$. This way, the augmented samples are diverse by emphasizing different frequency components via random weights, and at the same time coherent with original distributions as they contain the same basic sets of components.

\subsection{Time-Domain Augmentation}
Complementing the frequency-domain information, time domain also provides useful patterns. Therefore, we propose to mix up sequences in the time domain, inspired by Mix-up augmentation \cite{zhang2017mixup}. 
Mix-up was originally designed for classification, and we adapt it to time-series forecasting by mixing up values at both past timestamps $1,\dots,d$ and future timestamps $d+1,\dots,T$. 
Assume $\mathbf{S_i}=[\mathbf{H_i},\mathbf{F_i}]$ and $\mathbf{S_j}=[\mathbf{H_j},\mathbf{F_j}]$ are two randomly sampled sequences after spectral augmentation, where $\mathbf{H_i}=[\mathbf{s_i^1},\dots,\mathbf{s_i^d}],\mathbf{F_i}=[\mathbf{s_i^{d+1}},\dots,\mathbf{s_i^T}]$ respectively represents past and future data, similarly for $\mathbf{S_j}$. We construct new sequence as $\mathbf{S'} = [\mathbf{H'},\mathbf{F'}]$, where
\begin{equation}
    \mathbf{H'} = \lambda \mathbf{H_i} + (1-\lambda) \mathbf{H_j},
\end{equation}
\begin{equation}
    \mathbf{F'} = \lambda \mathbf{F_i} + (1-\lambda) \mathbf{F_j},
\end{equation}
where $\lambda$ is sampled from a Beta distribution, i.e., $\lambda \sim \mathrm{Beta}(\alpha,\alpha)$, and $\alpha$ is the hyper-parameter that controls how similar the newly constructed sequence is compared with the original sequences $\mathbf{S_i}$ and $\mathbf{S_j}$. Mix-up augments patterns in the time domain meanwhile by its interpolation nature generates only linearly in-between coherent samples.

\subsection{Time-Series Forecasting}
For each training iteration, we apply both frequency-domain and time-domain augmentation to obtain an augmented series $\mathbf{S'}=[\mathbf{H'},\mathbf{F'}]$. During training, we feed the augmented series history $\mathbf{H'}$ as input, and optimize the forecasting model through reconstructing future part of the series $\mathbf{F'}$. In our experiments, we adopt the state-of-the-art forecasting model Informer~\cite{zhou2021informer} (AAAI 2021 best paper) as the base forecaster. To minimize the reconstruction loss $\mathcal{L}$, we calculate Mean Square Error (MSE) between the forecasting model output $\mathbf{Y}$ and the ground-truth future part $\mathbf{F'}$: 
\begin{equation}
    \mathcal{L} = \frac{1}{N}\sum_{i=1}^N ||\mathbf{Y_i}-\mathbf{F'_i}||_2^2,
\end{equation}
where $N$ is the number of augmented series in training set.

\begin{table*}[t]
\centering
\begin{footnotesize}
\caption{MSE and MAE on benchmark datasets with input context length $96$ and forecasting horizon $ \{96,192,336,720\}$. We \textbf{bold} the best performing results, \underline{underline} the second best, and \dashuline{mark with dashline} the best baseline.}
\scalebox{0.77}{
\begin{tabular}{c|c|cccccccccccccc|cccccc}
\toprule
\multicolumn{2}{c|}{Methods}&\multicolumn{2}{c|}{None}&\multicolumn{2}{c|}{WW}&\multicolumn{2}{c|}{RobustTAD}&\multicolumn{2}{c|}{STL}&\multicolumn{2}{c|}{EMD-R}&\multicolumn{2}{c|}{GAN}&\multicolumn{2}{c|}{DBA}&\multicolumn{2}{c|}{\timeabl}&\multicolumn{2}{c|}{\freqabl}&\multicolumn{2}{c}{\our}\\
\midrule
\multicolumn{2}{c|}{Metric} & MSE & MAE & MSE & MAE & MSE & MAE & MSE & MAE & MSE & MAE & MSE & MAE & MSE & MAE & MSE & MAE & MSE & MAE & MSE & MAE \\
\midrule
\multirow{4}{*}{\rotatebox{90}{ETTh1}} & 96 & 0.947 & 0.760 &0.894&0.730&1.011&0.789&0.910&0.738&\dashuline{0.838}&\dashuline{0.686}&0.904&0.734&0.943&0.759& \underline{0.658}&\underline{0.573}&0.841&0.693&\textbf{0.645}&\textbf{0.572}\\ 
& 192 &0.977&0.767&0.999&0.775&0.946&0.744&\dashuline{0.942}&0.740&\dashuline{0.942}&\dashuline{0.729}&0.985&0.757&0.977&0.767&\textbf{0.754}&\textbf{0.616}&0.962&0.755&\underline{0.772}&\underline{0.628} \\
& 336 & 1.112&0.833&1.058 &0.808 &1.094&0.816&1.086&0.814&\dashuline{1.056}&\dashuline{0.804}&1.070&0.816&1.111&0.832&\textbf{0.860}&\textbf{0.681}&1.103&0.829&\underline{0.889}&\underline{0.705}\\
& 720 & 1.182&0.864&1.201 &0.884 &\dashuline{1.146}&\dashuline{0.841}&1.187 &0.863 &1.187&0.866&1.221&0.892&1.184&0.866&\underline{0.994}&\underline{0.752}&1.162&0.857&\textbf{0.972}&\textbf{0.746} \\
\midrule
\multirow{4}{*}{\rotatebox{90}{ETTh2}} &96&{3.084}&{1.383}&3.399 &1.463 &3.028&1.380&3.042&1.374&3.440&1.451&\dashuline{2.906}&1.377&3.005&\dashuline{1.370}&\textbf{1.802}&\textbf{1.012}&2.984&1.371&\underline{1.869}&\underline{1.046} \\ 
& 192&5.966&2.023& 6.432& 2.098& 6.203&2.083&5.894&2.023&6.326&2.089&\dashuline{5.467}&\dashuline{1.939}&5.636&\dashuline{1.939}&\underline{4.023}&\underline{1.569}&5.229&1.886&\textbf{3.467}&\textbf{1.483} \\
& 336&4.775&\dashuline{1.829}&4.965 &1.852 &5.287&1.883&4.774&1.833&5.436&1.948&4.906&1.868&\dashuline{4.765}&\dashuline{1.829}&\underline{3.391}&\textbf{1.482}&4.374&1.736&\textbf{3.365}&\underline{1.513} \\
& 720&3.991&1.694&3.828 &1.677 & 3.896&1.659&\dashuline{3.563}&\dashuline{1.579}&4.082&1.717&4.060&1.744&4.009&1.698&\underline{2.700}&\underline{1.377}&3.571&1.619&\textbf{2.621}&\textbf{1.340}\\
\midrule
\multirow{4}{*}{\rotatebox{90}{ETTm1}} &96 & 0.633&0.570&0.629&0.570&0.593&0.546&\dashuline{0.549}&\dashuline{0.534}&0.640&0.564&0.626&0.576&0.634&0.573&\underline{0.428}&\underline{0.439}&0.550&0.525&\textbf{0.403}&\textbf{0.424}\\ 
& 192 & 0.797&0.673&0.792&0.665&0.802&0.663&0.738&0.646&\dashuline{0.682}&\dashuline{0.590}&0.842&0.689&0.803&0.675&\underline{0.538}&\underline{0.505}&{0.668}&{0.591}&\textbf{0.489}&\textbf{0.482}\\
& 336 &1.149&0.839&0.997 &0.774 &0.983&0.759&0.992&0.773&\dashuline{0.869}&\dashuline{0.695}&0.902&0.725&1.146&0.838&\underline{0.648}&\underline{0.567}&0.900&0.715&\textbf{0.627}&\textbf{0.559} \\
& 720 & 1.128&0.805&1.077 &0.786 &1.063&0.789&1.184 &0.820 &\dashuline{1.008}&\dashuline{0.748}&1.225&0.862&1.131&0.806&\underline{0.828}&\underline{0.657}&0.934&0.720&\textbf{0.764}&\textbf{0.628}\\
\midrule
\multirow{4}{*}{\rotatebox{90}{ETTm2}} &96 &0.415&0.506&0.385&0.476&0.386&0.476&{0.388}&{0.489}&0.398&0.478&\dashuline{0.334}&\dashuline{0.430}&0.415&0.506&\underline{0.297}&\underline{0.394}&{0.328}&{0.423}&\textbf{0.292}&\textbf{0.388} \\ 
& 192 &0.776&0.675&0.784&0.680&0.875&0.724&0.712&0.650&\dashuline{0.542}&\dashuline{0.551}&0.625&0.604&0.767&0.673&\textbf{0.420}&\textbf{0.490}&0.592&0.585&\underline{0.429}&\underline{0.497} \\
& 336 &1.515&0.946&1.425 &0.923 &1.390&0.909&1.402&0.908&\dashuline{1.287}&\dashuline{0.875}&1.453&0.929&1.568&0.962&\underline{0.943}&\underline{0.735}&{1.287}&{0.876}&\textbf{0.811}&\textbf{0.692} \\
& 720 & 3.462&1.423&4.370 &1.637 &4.792&1.654& 3.271& 1.378&3.247 &\dashuline{1.353}&\dashuline{3.129}&1.355&3.475&1.427 &\textbf{2.645}&\textbf{1.201}&3.765&1.509&\underline{2.790}&\underline{1.268}\\
\midrule
\multirow{4}{*}{\rotatebox{90}{Exchange}} &96 &0.959&0.784&\dashuline{0.799}&\dashuline{0.687}&0.962&0.788&0.860&0.746&0.923&0.761&0.942&0.772&0.958&0.784&\underline{0.285}&\underline{0.402}&0.809&0.708&\textbf{0.271}&\textbf{0.395} \\ 
& 192 &1.113&0.837&\dashuline{1.084} &\dashuline{0.787} &1.180&0.873&1.143&0.852&1.182&0.853&1.115&0.855&1.109 &0.836 &\underline{0.627}&\underline{0.615}&1.125&0.822&\textbf{0.606}&\textbf{0.605} \\
& 336 &1.605&1.009&1.697 &0.995 &1.562&1.006&1.548&0.987&1.606&1.001&\dashuline{1.465}&\dashuline{0.972}&1.595 &1.007 &\underline{1.007}&\underline{0.807}&1.438&0.944&\textbf{0.935}&\textbf{0.779} \\
& 720 &2.847&1.390&3.198 &1.481 &2.816&1.380&\dashuline{2.484} &\dashuline{1.297} &2.914&1.413&2.571&1.309&2.847&1.390 &\textbf{1.409}&\textbf{0.928}&{2.056}&1.153&\underline{1.771}&\underline{1.052} \\
\bottomrule
\end{tabular}
\label{tab:main}
}
\end{footnotesize}
\end{table*}

\section{Experimental Setup and Results}
\label{sec:exp}

\subsection{Datasets, Baselines and Experimental Setup}
We evaluate our augmentation method \our on five time-series datasets: ETTh1, ETTh2, ETTm1, ETTm2, Exchange~\cite{zhou2021informer,wu2021autoformer}. We follow previous studies~\cite{zhou2021informer,wu2021autoformer} for $96$ context length and $96, 192, 336, 720$ forecasting horizon. All the datasets in our experiments are multivariate, and we apply EMD and mix-up separately for each original variable, and concatenate them to obtain the augmented multivariate time series. We follow previous works for $7:2:1$ train, validation and test set split, and evaluate the performance with Mean Square Error (MSE) and Mean Absolute Error (MAE). 

We use Informer~\cite{zhou2021informer} as the base forecasting model. $\alpha$ of the $\mathrm{Beta}$ distribution equals $0.5$, which is chosen from a grid search of $\{0.25,0.5,0.75,1.0\}$. $a, b$ of the Uniform distribution $U(a,b)$ are set to $0$ and $2$, respectively. We use Adam optimizer with a decaying learning rate starting from $1\mathrm{e}^{-4}$. 
We compare \our with base model without any augmentation (None), as well as state-of-the-art time series augmentation methods: WW~\cite{le2016data}, DBA~\cite{forestier2017generating,fawaz2018data,bandara2021improving}, EMD-R~\cite{nam2020data}, GAN~\cite{esteban2017real}, STL~\cite{bergmeir2016bagging,bandara2021improving}, RobustTAD~\cite{gao2020robusttad}. We conducted careful grid search for hyper-parameter tuning for each baseline. We repeat all the experiments for $3$ runs and record both average performance and standard deviation. 

\subsection{Main Results}

\begin{figure}[t]
      \centering
      \begin{subfigure}[b]{0.235\textwidth}
          \centering
          \includegraphics[width=\textwidth]{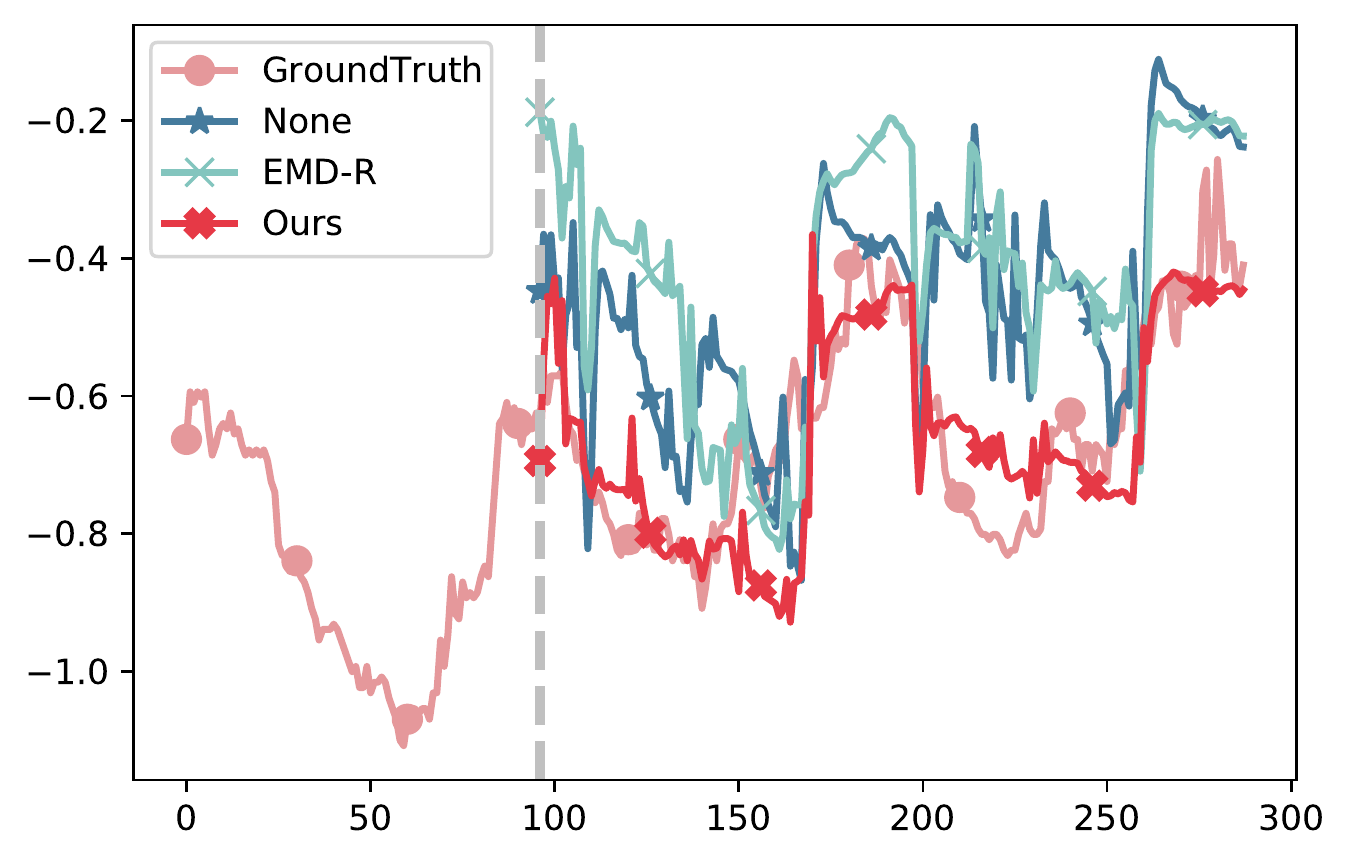}
          \caption{ETTm1}
          \label{fig:case-ettm1}
      \end{subfigure}
      \begin{subfigure}[b]{0.235\textwidth}
          \centering
          \includegraphics[width=\textwidth]{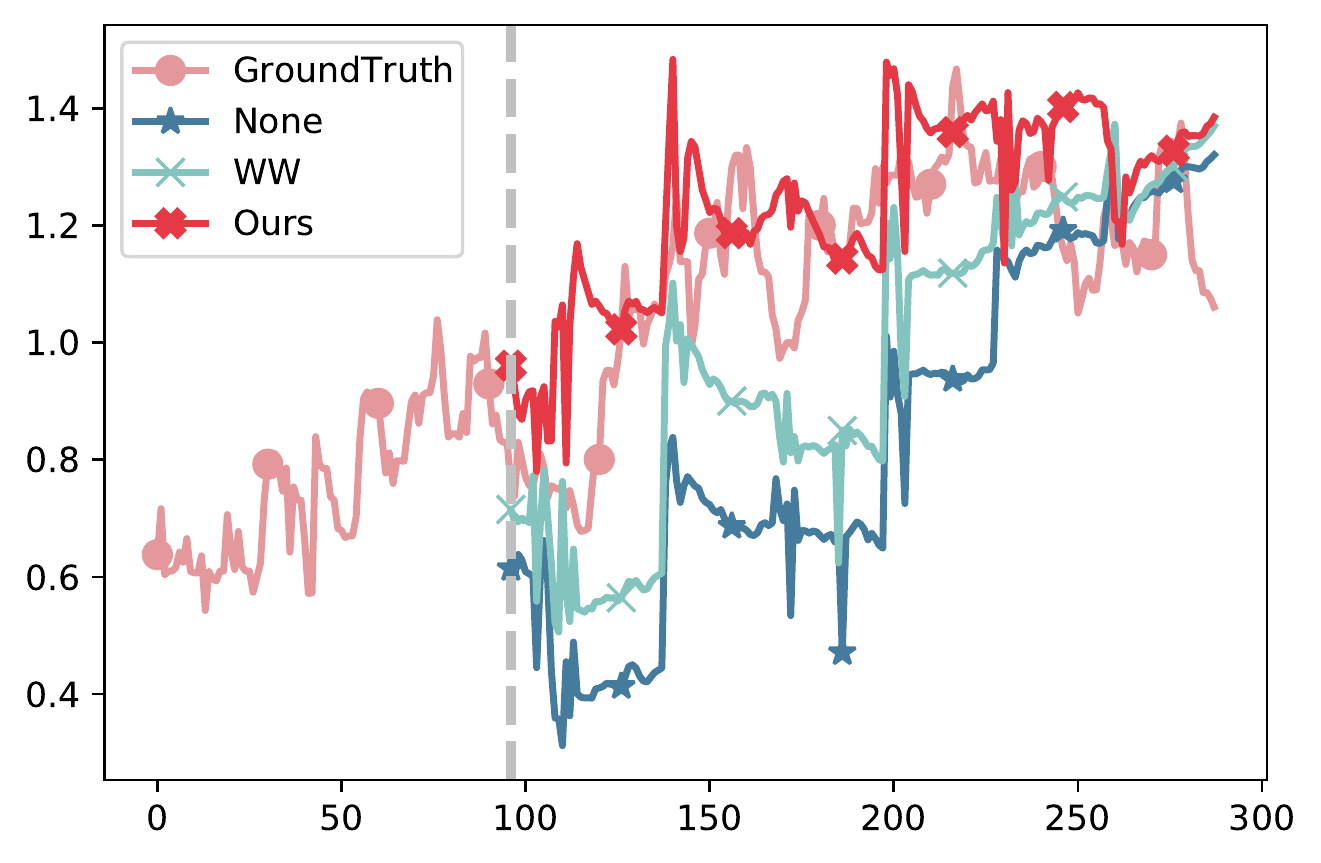}
          \caption{Exchange}
          \label{fig:case-ex}
      \end{subfigure}
         \caption{Predictions on ETTm1 and Exchange with $192$ horizon. \our predicts data that best align with the ground truth.} 
         \label{fig:pred}
\end{figure}

We evaluate \our and baselines, and report the average performance in Table \ref{tab:main}. Due to space limit, we report the full results (including both average performance and standard deviation) in Table~\ref{tab:main-std} in Appendix. For easier comparison, we also scale MSE (average and standard deviation) with respect to the average MSE of the base model without augmentation. Similarly, we scaled MAE with respect to the average MAE of the base model without augmentation. Results are shown in Table~\ref{tab:main-std-scale} in Appendix. We bold the best results,  underline the second best, and mark with dash line the best baseline. \our consistently outperforms baselines on different datasets by jointly leveraging time-domain and frequency-domain information, with an average reduction of 28.2\% for MSE and 18.0\% for MAE, compared with the best baseline for each dataset. 
To evaluate the statistical significance, we also run the Friedman test and the Wilcoxon-signed rank test with Holm’s $\alpha$ (5\%) following previous work~\cite{li2021shapenet}. The Friedman test shows statistical significance $p = 0.00$ (much smaller than $\alpha = 0.05$), so there exists a significant difference among different methods. The Wilcoxon-signed rank test indicates the statistical significance of STAug compared with all the baselines with $p = 0.00$ far below 0.05.
We also present qualitative comparisons in Figure~\ref{fig:pred}. Compared with the predictions by the base model and the best-performing baseline for the corresponding setting, \our forecasts values that better align with the original time series. 

\begin{figure}[t]
\centering
\includegraphics[width=1\linewidth]{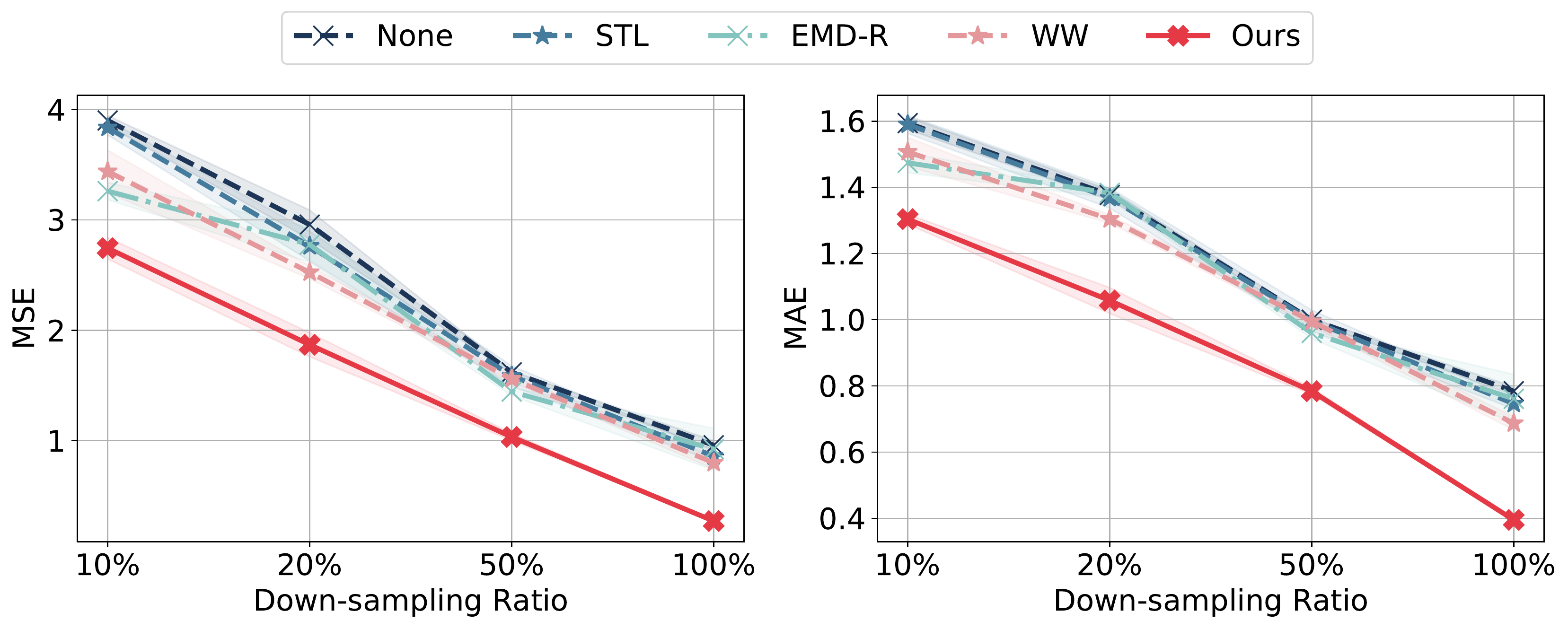}
\caption{MSE and MAE for different down-sampling ratios.
}
\label{fig:ds}

\end{figure}

\subsection{Ablation Study}
To examine the effect of augmentation in different domains, we also conduct ablation study by augmenting data only in the frequency or time domain in Table~\ref{tab:main}. \freqabl stands for the model that \emph{removes} frequency-domain augmentation, and \timeabl stands for the model that \emph{removes} time-domain data augmentation. The performance degrades after removing augmentation in either the frequency or time domain, validating the need of combining both domains. 

\subsection{Robustness Study}

\label{sec:robust}

We sub-sample 10\%, 20\%, 50\% of the Exchange dataset, and compare \our with the base model and the best-performing baseline methods for the corresponding setting (horizon $96$). We report on MSE and MAE in Figure~\ref{fig:ds}. The performances of different methods increase as we have more data. When the sample size is small, the performance gap between augmentation methods and base model without augmentation becomes larger, which shows that augmentation is especially helpful when the original dataset is small. \our leverages information from both time and frequency domains, and performs consistently better with respect to the original sample size. Moreover, \our shows more significant improvement over base model and the best-performing baselines with fewer available samples in the original dataset, which demonstrates its robustness with respect to the number of data samples.

\vspace{-0.5em}
\section{Conclusion}
\vspace{-0.5em}
We proposed a generic yet effective time-series data augmentation method \our to combine patterns in both the time domain and frequency domain. In the frequency domain, the re-combined subcompacts of the original time series are both diverse and preserve the original basic components. In the time domain, we adapt the mix-up strategy to generate diverse and in-between coherent samples by linearly interpolating both past and future part of a time series. Experiments on five real-world time-series datasets show that \our best reduces the forecasting errors of base model compared with existing augmentation methods. We also performed robustness analysis and observed that \our stays robust with respect to sampling size. 
Our ongoing work in the area involves systematically studying the effectiveness of various time domain and frequency domain augmentation methods, and designing data-dependent selection procedure to choose the most suitable augmentation method for different datasets.

\small
\bibliographystyle{IEEEbib}
\bibliography{refs-reduced}

\section{Appendix}

\begin{table*}[hb]
\centering
\begin{footnotesize}
\caption{MSE, MAE (average and standard deviation) with input context length $96$ and forecasting horizon $ \{96,192,336,720\}$. We \textbf{bold} the best performing results, \underline{underline} the second best, and \dashuline{mark with dashline} the best baseline.}
\scalebox{0.52}{
\begin{tabular}{c|c|cccccccccccccc|cccccc}
\toprule
\multicolumn{2}{c|}{Methods}&\multicolumn{2}{c|}{None}&\multicolumn{2}{c|}{WW}&\multicolumn{2}{c|}{RobustTAD}&\multicolumn{2}{c|}{STL}&\multicolumn{2}{c|}{EMD-R}&\multicolumn{2}{c|}{GAN}&\multicolumn{2}{c|}{DBA}&\multicolumn{2}{c|}{\timeabl}&\multicolumn{2}{c|}{\freqabl}&\multicolumn{2}{c}{\our}\\
\midrule
\multicolumn{2}{c|}{Metric} & MSE & MAE & MSE & MAE & MSE & MAE & MSE & MAE & MSE & MAE & MSE & MAE & MSE & MAE & MSE & MAE & MSE & MAE & MSE & MAE \\
\midrule
\multirow{4}{*}{\rotatebox{90}{ETTh1}} & 96 & 0.95$\pm$0.06 & 0.76$\pm$0.04 & 0.89$\pm$0.06 & 0.73$\pm$0.04 & 1.01$\pm$0.03 &0.79$\pm$0.01 & 0.91$\pm$0.05 & 0.74$\pm$0.03 & \dashuline{0.84$\pm$0.05} & \dashuline{0.69$\pm$0.03} & 0.90$\pm$0.08 & 0.73$\pm$0.06 & 0.94$\pm$0.06 & 0.76$\pm$0.04 & \underline{0.66$\pm$0.01}&\underline{0.57$\pm$0.01}& 0.84$\pm$0.06 & 0.69$\pm$0.03 & \textbf{0.65$\pm$0.01} & \textbf{0.57$\pm$0.01}\\ 
& 192 & 0.98$\pm$0.02 & 0.77$\pm$0.02 & 1.00$\pm$0.12 & 0.78$\pm$0.08 & 0.95$\pm$0.02 & 0.74$\pm$0.01 & \dashuline{0.94$\pm$0.02} & 0.74$\pm$0.02 & \dashuline{0.94$\pm$0.04} & \dashuline{0.73$\pm$0.04} & 0.99$\pm$0.04 & 0.76$\pm$0.03 & 0.98$\pm$0.02 & 0.77$\pm$0.02 & \textbf{0.75$\pm$0.02} & \textbf{0.62$\pm$0.01} & 0.96$\pm$0.00 & 0.76$\pm$0.01 & \underline{0.77$\pm$0.03} & \underline{0.63$\pm$0.02} \\
& 336 & 1.11$\pm$0.04 & 0.83$\pm$0.02 & 1.06$\pm$0.03 & 0.81$\pm$0.02 & 1.09$\pm$0.06 & 0.82$\pm$0.04 & 1.09$\pm$0.04 & 0.81$\pm$0.02 & \dashuline{1.06$\pm$0.02} & \dashuline{0.80$\pm$0.01} & 1.07$\pm$0.01 & 0.82$\pm$0.01 & 1.11$\pm$0.04 & 0.83$\pm$0.02 & \textbf{0.86$\pm$0.02} & \textbf{0.68$\pm$0.01} & 1.10$\pm$0.06 & 0.83$\pm$0.03 & \underline{0.89$\pm$0.02} & \underline{0.71$\pm$0.01}\\
& 720 & 1.18$\pm$0.03 & 0.86$\pm$0.02 & 1.20$\pm$0.02 & 0.88$\pm$0.01 & \dashuline{1.15$\pm$0.02} & \dashuline{0.84$\pm$0.01} & 1.19$\pm$0.02 & 0.86$\pm$0.01 & 1.19$\pm$0.02 & 0.87$\pm$0.01 & 1.22$\pm$0.04 & 0.89$\pm$0.01 & 1.18$\pm$0.03 & 0.87$\pm$0.02 & \underline{0.99$\pm$0.06} & \underline{0.75$\pm$0.03} & 1.16$\pm$0.01 & 0.86$\pm$0.01 & \textbf{0.97$\pm$0.00} & \textbf{0.75$\pm$0.00} \\
\midrule
\multirow{4}{*}{\rotatebox{90}{ETTh2}} &96&{3.08$\pm$0.62} & {1.38$\pm$0.13} & 3.40$\pm$0.48 & 1.46$\pm$0.11 & 3.03$\pm$0.25 & 1.38$\pm$0.05 & 3.04$\pm$0.48 & 1.37$\pm$0.11 & 3.44$\pm$0.29 & 1.45$\pm$0.05 & \dashuline{2.91$\pm$0.36} & 1.38$\pm$0.09 & 3.01$\pm$0.54 & \dashuline{1.37$\pm$0.12} & \textbf{1.80$\pm$0.11} & \textbf{1.01$\pm$0.04} & 2.98$\pm$0.57 & 1.37$\pm$0.14 & \underline{1.87$\pm$0.30} & \underline{1.05$\pm$0.09} \\ 
& 192& 5.97$\pm$0.30 & 2.02$\pm$0.05 & 6.43$\pm$0.68 & 2.10$\pm$0.09 & 6.20$\pm$0.35 & 2.08$\pm$0.05 & 5.89$\pm$0.46 & 2.02$\pm$0.08 & 6.33$\pm$0.10 & 2.09$\pm$0.00 & \dashuline{5.47$\pm$0.30} & \dashuline{1.94$\pm$0.05} & 5.64$\pm$0.47 & \dashuline{1.94$\pm$0.09} & \underline{4.02$\pm$0.23} & \underline{1.57$\pm$0.05} & 5.23$\pm$0.15 & 1.89$\pm$0.03 & \textbf{3.47$\pm$0.24} & \textbf{1.48$\pm$0.06} \\
& 336 & 4.78$\pm$0.33& \dashuline{1.83$\pm$0.08} & 4.97$\pm$0.07 & 1.85$\pm$0.05 & 5.29$\pm$0.15 & 1.88$\pm$0.01 & 4.77$\pm$0.24 & 1.83$\pm$0.05 & 5.44$\pm$0.18 & 1.95$\pm$0.05 & 4.91$\pm$0.31 & 1.87$\pm$0.06 & \dashuline{4.77$\pm$0.32} & \dashuline{1.83$\pm$0.08} & \underline{3.39$\pm$0.27} & \textbf{1.48$\pm$0.07} & 4.37$\pm$0.26 & 1.74$\pm$0.05 & \textbf{3.37$\pm$0.14} & \underline{1.51$\pm$0.04} \\
& 720& 3.99$\pm$0.27 & 1.69$\pm$0.06 & 3.83$\pm$0.13 & 1.68$\pm$0.04 & 3.90$\pm$0.13 & 1.66$\pm$0.00 & \dashuline{3.56$\pm$0.04} & \dashuline{1.58$\pm$0.03} & 4.08$\pm$0.09 & 1.72$\pm$0.03 & 4.06$\pm$0.41 & 1.74$\pm$0.10 & 4.01$\pm$0.27 & 1.70$\pm$0.06 & \underline{2.70$\pm$0.30} & \underline{1.38$\pm$0.09} & 3.57$\pm$0.18 & 1.62$\pm$0.04 & \textbf{2.62$\pm$0.17} & \textbf{1.34$\pm$0.08}\\
\midrule
\multirow{4}{*}{\rotatebox{90}{ETTm1}} &96 & 0.63$\pm$0.03 & 0.57$\pm$0.01 & 0.63$\pm$0.02 & 0.57$\pm$0.01 & 0.59$\pm$0.03 & 0.55$\pm$0.02 & \dashuline{0.55$\pm$0.03} & \dashuline{0.53$\pm$0.01} & 0.64$\pm$0.01 & 0.56$\pm$0.00 & 0.63$\pm$0.02 & 0.58$\pm$0.01 & 0.63$\pm$0.03 & 0.57$\pm$0.02 & \underline{0.43$\pm$0.00} & \underline{0.44$\pm$0.00} & 0.55$\pm$0.03 & 0.53$\pm$0.02 & \textbf{0.40$\pm$0.00} & \textbf{0.42$\pm$0.00}\\ 
& 192 & 0.80$\pm$0.04 & 0.67$\pm$0.02 & 0.79$\pm$0.07 & 0.67$\pm$0.03 & 0.80$\pm$0.08 & 0.66$\pm$0.04 & 0.74$\pm$0.06 & 0.65$\pm$0.03 & \dashuline{0.68$\pm$0.01} & \dashuline{0.59$\pm$0.01} & 0.84$\pm$0.06 & 0.69$\pm$0.02 & 0.80$\pm$0.03 & 0.68$\pm$0.02 & \underline{0.54$\pm$0.02} & \underline{0.51$\pm$0.00} & {0.67$\pm$0.04} & {0.59$\pm$0.02} & \textbf{0.49$\pm$0.00} & \textbf{0.48$\pm$0.00}\\
& 336 &1.15$\pm$0.07 & 0.84$\pm$0.03 & 1.00$\pm$0.06 & 0.77$\pm$0.03 & 0.98$\pm$0.04 & 0.76$\pm$0.03 & 0.99$\pm$0.07 & 0.77$\pm$0.03 & \dashuline{0.87$\pm$0.14} & \dashuline{0.70$\pm$0.06} & 0.90$\pm$0.07 & 0.73$\pm$0.04 & 1.15$\pm$0.06 & 0.84$\pm$0.02 & \underline{0.65$\pm$0.03} & \underline{0.57$\pm$0.02} & 0.90$\pm$0.02 & 0.72$\pm$0.01 & \textbf{0.63$\pm$0.03} & \textbf{0.56$\pm$0.01} \\
& 720 & 1.13$\pm$0.04 & 0.81$\pm$0.00 & 1.08$\pm$0.07 & 0.79$\pm$0.02 & 1.06$\pm$0.04 & 0.79$\pm$0.03 & 1.18$\pm$0.02 & 0.82$\pm$0.00 & \dashuline{1.01$\pm$0.10} & \dashuline{0.75$\pm$0.05} & 1.23$\pm$0.12 & 0.86$\pm$0.04 & 1.13$\pm$0.04 & 0.81$\pm$0.01 & \underline{0.83$\pm$0.03} & \underline{0.66$\pm$0.01} & 0.93$\pm$0.10 & 0.72$\pm$0.04 & \textbf{0.76$\pm$0.04}&\textbf{0.63$\pm$0.02}\\
\midrule
\multirow{4}{*}{\rotatebox{90}{ETTm2}} &96 &0.42$\pm$0.06 & 0.51$\pm$0.04 & 0.39$\pm$0.02 & 0.48$\pm$0.02 & 0.39$\pm$0.02 & 0.48$\pm$0.01 & {0.39$\pm$0.03} & {0.49$\pm$0.03} & 0.40$\pm$0.08 & 0.48$\pm$0.05 & \dashuline{0.33$\pm$0.03} & \dashuline{0.43$\pm$0.03} & 0.42$\pm$0.05 & 0.51$\pm$0.04 & \underline{0.30$\pm$0.03} & \underline{0.39$\pm$0.01} & {0.33$\pm$0.06} & {0.42$\pm$0.04} & \textbf{0.29$\pm$0.03}&\textbf{0.39$\pm$0.03} \\ 
& 192 &0.78$\pm$0.09 & 0.68$\pm$0.03 & 0.78$\pm$0.09 & 0.68$\pm$0.04 & 0.88$\pm$0.20 & 0.72$\pm$0.10 & 0.71$\pm$0.05 & 0.65$\pm$0.01 & \dashuline{0.54$\pm$0.04} & \dashuline{0.55$\pm$0.03} & 0.63$\pm$0.11 & 0.60$\pm$0.05 & 0.77$\pm$0.07 & 0.67$\pm$0.03 & \textbf{0.42$\pm$0.00} & \textbf{0.49$\pm$0.01} & 0.59$\pm$0.01 & 0.59$\pm$0.01 & \underline{0.43$\pm$0.02}&\underline{0.50$\pm$0.01} \\
& 336 &1.52$\pm$0.07 & 0.95$\pm$0.02 & 1.43$\pm$0.08 & 0.92$\pm$0.02 & 1.39$\pm$0.18& 0.91$\pm$0.07 & 1.40$\pm$0.06 & 0.91$\pm$0.03 & \dashuline{1.29$\pm$0.04} &\dashuline{0.88$\pm$0.02} & 1.45$\pm$0.12 & 0.93$\pm$0.05 & 1.57$\pm$0.10 & 0.96$\pm$0.03 & \underline{0.94$\pm$0.19}&\underline{0.74$\pm$0.07} &{1.29$\pm$0.11} &{0.88$\pm$0.03} &\textbf{0.81$\pm$0.03}&\textbf{0.69$\pm$0.01} \\
& 720 & 3.46$\pm$0.27& 1.42$\pm$0.08 & 4.37$\pm$0.29 & 1.64$\pm$0.08 & 4.79$\pm$0.03 &1.65$\pm$0.01 & 3.27$\pm$0.30 & 1.38$\pm$0.08 & 3.25$\pm$0.35 &\dashuline{1.35$\pm$0.07} &\dashuline{3.13$\pm$0.71} &1.36$\pm$0.17 & 3.48$\pm$0.27 & 1.43$\pm$0.08 &\textbf{2.65$\pm$0.47} &\textbf{1.20$\pm$0.10} & 3.77$\pm$0.38 & 1.51$\pm$0.10 & \underline{2.79$\pm$0.25}&\underline{1.27$\pm$0.04}\\
\midrule
\multirow{4}{*}{\rotatebox{90}{Exchange}} &96 &0.96$\pm$0.04 & 0.78$\pm$0.01 & \dashuline{0.80$\pm$0.06} &\dashuline{0.69$\pm$0.02} &0.96$\pm$0.04 & 0.79$\pm$0.02 & 0.86$\pm$0.05 & 0.75$\pm$0.02 & 0.92$\pm$0.19 & 0.76$\pm$0.08 & 0.94$\pm$0.06 & 0.77$\pm$0.03 & 0.96$\pm$0.04 & 0.78$\pm$0.01 & \underline{0.29$\pm$0.01} &\underline{0.40$\pm$0.00} & 0.81$\pm$0.02 & 0.71$\pm$0.01 & \textbf{0.27$\pm$0.00} &\textbf{0.40$\pm$0.01} \\ 
& 192 &1.11$\pm$0.01 & 0.84$\pm$0.00 & \dashuline{1.08$\pm$0.00} & \dashuline{0.79$\pm$0.00} & 1.18$\pm$0.01 & 0.87$\pm$0.01 & 1.14$\pm$0.04 & 0.85$\pm$0.01 & 1.18$\pm$0.02 & 0.85$\pm$0.01 & 1.12$\pm$0.02 & 0.86$\pm$0.01 & 1.11$\pm$0.01 & 0.84$\pm$0.00 &\underline{0.63$\pm$0.05} &\underline{0.62$\pm$0.03} & 1.13$\pm$0.05 & 0.82$\pm$0.01& \textbf{0.61$\pm$0.04} &\textbf{0.61$\pm$0.02} \\
& 336 &1.61$\pm$0.04 & 1.01$\pm$0.02 & 1.70$\pm$0.07 & 1.00$\pm$0.02 & 1.56$\pm$0.03 & 1.01$\pm$0.00 & 1.55$\pm$0.02 & 0.99$\pm$0.01 & 1.61$\pm$0.09 & 1.00$\pm$0.04 & \dashuline{1.47$\pm$0.11} & \dashuline{0.97$\pm$0.03} & 1.60$\pm$0.04 & 1.01$\pm$0.02 &\underline{1.01$\pm$0.13} &\underline{0.81$\pm$0.05} &1.44$\pm$0.02 & 0.94$\pm$0.01 & \textbf{0.94$\pm$0.12}&\textbf{0.78$\pm$0.05} \\
& 720 & 2.85$\pm$0.16 & 1.39$\pm$0.04 & 3.20$\pm$0.08  & 1.48$\pm$0.02 & 2.82$\pm$0.20 & 1.38$\pm$0.07 & \dashuline{2.48$\pm$0.22} & \dashuline{1.30$\pm$0.06} & 2.91$\pm$0.04 & 1.41$\pm$0.01 & 2.57$\pm$0.17 & 1.31$\pm$0.07 & 2.85$\pm$0.16 & 1.39$\pm$0.04 &\textbf{1.41$\pm$0.24} & \textbf{0.93$\pm$0.08} & {2.06$\pm$0.16} & 1.15$\pm$0.05 & \underline{1.77$\pm$0.13}&\underline{1.05$\pm$0.07} \\
\bottomrule
\end{tabular}
\label{tab:main-std}
}
\end{footnotesize}
\end{table*}

\begin{table*}[hb]
\centering
\begin{footnotesize}
\caption{Scaled MSE, MAE (average, standard deviation) with context length $96$ and forecasting horizon $\{96,192,336,720\}$. We \textbf{bold} the best performing results, \underline{underline} the second best, and \dashuline{mark with dashline} the best baseline.}
\scalebox{0.52}{
\begin{tabular}{c|c|cccccccccccccc|cccccc}
\toprule
\multicolumn{2}{c|}{Methods}&\multicolumn{2}{c|}{None}&\multicolumn{2}{c|}{WW}&\multicolumn{2}{c|}{RobustTAD}&\multicolumn{2}{c|}{STL}&\multicolumn{2}{c|}{EMD-R}&\multicolumn{2}{c|}{GAN}&\multicolumn{2}{c|}{DBA}&\multicolumn{2}{c|}{\timeabl}&\multicolumn{2}{c|}{\freqabl}&\multicolumn{2}{c}{\our}\\
\midrule
\multicolumn{2}{c|}{Metric} & MSE & MAE & MSE & MAE & MSE & MAE & MSE & MAE & MSE & MAE & MSE & MAE & MSE & MAE & MSE & MAE & MSE & MAE & MSE & MAE \\
\midrule
\multirow{4}{*}{\rotatebox{90}{ETTh1}} & 96 & 1.00$\pm$0.07 & 1.00$\pm$0.05 & 0.94$\pm$0.07 & 0.96$\pm$0.05 & 1.07$\pm$0.03 &1.04$\pm$0.02 & 0.96$\pm$0.05 & 0.97$\pm$0.04& \dashuline{0.89$\pm$0.05} & \dashuline{0.90$\pm$0.04} & 0.96$\pm$0.09 & 0.97$\pm$0.08 & 1.00$\pm$0.06 & 1.00$\pm$0.05 & \underline{0.69$\pm$0.02}&\underline{0.75$\pm$0.01}& 0.89$\pm$0.06 & 0.91$\pm$0.04 & \textbf{0.68$\pm$0.01} & \textbf{0.75$\pm$0.02}\\ 
& 192 &1.00$\pm$0.02 & 1.00$\pm$0.02 & 1.02$\pm$0.13 & 1.01$\pm$0.10 & 0.97$\pm$0.02 & 0.97$\pm$0.01 & \dashuline{0.96$\pm$0.02} & 0.96$\pm$0.02 & \dashuline{0.96$\pm$0.04} & \dashuline{0.95$\pm$0.05} & 1.01$\pm$0.04 & 0.99$\pm$0.04 & 1.00$\pm$0.02 & 1.00$\pm$0.02 & \textbf{0.77$\pm$0.02} & \textbf{0.80$\pm$0.02} & 0.99$\pm$0.00 & 0.99$\pm$0.01 & \underline{0.79$\pm$0.03} & \underline{0.82$\pm$0.02} \\
& 336 & 1.00$\pm$0.03 & 1.00$\pm$0.02 & 0.95$\pm$0.03 & 0.97$\pm$0.02 & 0.98$\pm$0.06 & 0.98$\pm$0.05 & 0.98$\pm$0.03 & 0.98$\pm$0.02 & \dashuline{0.95$\pm$0.02} & \dashuline{0.97$\pm$0.02} & 0.96$\pm$0.01 & 0.98$\pm$0.01 & 1.00$\pm$0.03 & 1.00$\pm$0.02 & \textbf{0.77$\pm$0.02} & \textbf{0.82$\pm$0.01} & 0.99$\pm$0.05 & 1.00$\pm$0.03 & \underline{0.80$\pm$0.02} & \underline{0.85$\pm$0.02}\\
& 720 & 1.00$\pm$0.02 & 1.00$\pm$0.02 & 1.02$\pm$0.02 & 1.02$\pm$0.01 & \dashuline{0.97$\pm$0.01} & \dashuline{0.97$\pm$0.01} & 1.00$\pm$0.01 & 1.00$\pm$0.02 & 1.00$\pm$0.01 & 1.00$\pm$0.01 & 1.03$\pm$0.03 & 1.03$\pm$0.01 & 1.00$\pm$0.02 & 1.00$\pm$0.02 & \underline{0.84$\pm$0.05} & \underline{0.87$\pm$0.03} & 0.98$\pm$0.01 & 0.99$\pm$0.01 & \textbf{0.82$\pm$0.00} & \textbf{0.86$\pm$0.00} \\
\midrule
\multirow{4}{*}{\rotatebox{90}{ETTh2}} &96&{1.00$\pm$0.20} & {1.00$\pm$0.09} & 1.10$\pm$0.15 & 1.06$\pm$0.08 & 0.98$\pm$0.08 & 1.00$\pm$0.03 & 0.99$\pm$0.16 & 0.99$\pm$0.08 & 1.12$\pm$0.09 & 1.05$\pm$0.03 & \dashuline{0.94$\pm$0.12} & 1.00$\pm$0.06 & 0.97$\pm$0.18 & \dashuline{0.99$\pm$0.09} & \textbf{0.58$\pm$0.03} & \textbf{0.73$\pm$0.03} & 0.97$\pm$0.19 & 0.99$\pm$0.10 & \underline{0.61$\pm$0.10} & \underline{0.76$\pm$0.06} \\ 
& 192& 1.00$\pm$0.05 & 1.00$\pm$0.02 & 1.08$\pm$0.11 & 1.04$\pm$0.04 & 1.04$\pm$0.06 & 1.03$\pm$0.03 & 0.99$\pm$0.08 & 1.00$\pm$0.04 & 1.06$\pm$0.02 & 1.03$\pm$0.00 & \dashuline{0.92$\pm$0.05} & \dashuline{0.96$\pm$0.02} & 0.95$\pm$0.08 & \dashuline{0.96$\pm$0.04} & \underline{0.67$\pm$0.04} & \underline{0.78$\pm$0.03} & 0.88$\pm$0.03 & 0.93$\pm$0.01 & \textbf{0.58$\pm$0.04} & \textbf{0.73$\pm$0.03} \\
& 336 & 1.00$\pm$0.07& \dashuline{1.00$\pm$0.04} & 1.04$\pm$0.01 & 1.01$\pm$0.03 & 1.11$\pm$0.03 & 1.03$\pm$0.00 & 1.00$\pm$0.05 & 1.00$\pm$0.03 & 1.14$\pm$0.04 & 1.07$\pm$0.03 & 1.03$\pm$0.07 & 1.02$\pm$0.04 & \dashuline{1.00$\pm$0.07} & \dashuline{1.00$\pm$0.04} & \underline{0.71$\pm$0.06} & \textbf{0.81$\pm$0.04} & 0.92$\pm$0.05 & 0.95$\pm$0.03 & \textbf{0.71$\pm$0.03} & \underline{0.83$\pm$0.02} \\
& 720& 1.00$\pm$0.07 & 1.00$\pm$0.04 & 0.96$\pm$0.03 & 0.99$\pm$0.02 & 0.98$\pm$0.03 & 0.98$\pm$0.00 & \dashuline{0.89$\pm$0.01} & \dashuline{0.93$\pm$0.02} & 1.02$\pm$0.02 & 1.01$\pm$0.02 & 1.02$\pm$0.10 & 1.03$\pm$0.06 & 1.00$\pm$0.07 & 1.00$\pm$0.04 & \underline{0.68$\pm$0.07} & \underline{0.81$\pm$0.05} & 0.90$\pm$0.05 & 0.96$\pm$0.03 & \textbf{0.66$\pm$0.04} & \textbf{0.79$\pm$0.05}\\
\midrule
\multirow{4}{*}{\rotatebox{90}{ETTm1}} &96 & 1.00$\pm$0.05 & 1.00$\pm$0.02 & 0.99$\pm$0.04 & 1.00$\pm$0.02 & 0.94$\pm$0.04 & 0.96$\pm$0.03 & \dashuline{0.87$\pm$0.04} & \dashuline{0.94$\pm$0.02} & 1.01$\pm$0.02 & 0.99$\pm$0.01 & 0.99$\pm$0.03 & 1.01$\pm$0.02 & 1.00$\pm$0.05 & 1.01$\pm$0.03 & \underline{0.68$\pm$0.00} & \underline{0.77$\pm$0.01} & 0.87$\pm$0.05 & 0.92$\pm$0.04 & \textbf{0.64$\pm$0.01} & \textbf{0.74$\pm$0.01}\\ 
& 192 & 1.00$\pm$0.04 & 1.00$\pm$0.03 & 0.99$\pm$0.09 & 0.99$\pm$0.04 & 1.01$\pm$0.11 & 0.99$\pm$0.06 & 0.93$\pm$0.07 & 0.96$\pm$0.04 & \dashuline{0.86$\pm$0.02} & \dashuline{0.88$\pm$0.02} & 1.06$\pm$0.07 & 1.02$\pm$0.04 & 1.01$\pm$0.04 & 1.00$\pm$0.03 & \underline{0.68$\pm$0.02} & \underline{0.75$\pm$0.00} & {0.84$\pm$0.05} & {0.88$\pm$0.03} & \textbf{0.61$\pm$0.01} & \textbf{0.72$\pm$0.00}\\
& 336 &1.00$\pm$0.06 & 1.00$\pm$0.04 & 0.87$\pm$0.05 & 0.92$\pm$0.03 & 0.86$\pm$0.04 & 0.90$\pm$0.03 & 0.86$\pm$0.06 & 0.92$\pm$0.04 & \dashuline{0.76$\pm$0.12} & \dashuline{0.83$\pm$0.08} & 0.79$\pm$0.06 & 0.86$\pm$0.04 & 1.00$\pm$0.05 & 1.00$\pm$0.03 & \underline{0.56$\pm$0.02} & \underline{0.68$\pm$0.02} & 0.78$\pm$0.02 & 0.85$\pm$0.01 & \textbf{0.55$\pm$0.02} & \textbf{0.67$\pm$0.01} \\
& 720 & 1.00$\pm$0.03 & 1.00$\pm$0.00 & 0.96$\pm$0.07 & 0.98$\pm$0.03 & 0.94$\pm$0.04 & 0.98$\pm$0.03 & 1.05$\pm$0.02 & 1.02$\pm$0.00 & \dashuline{0.89$\pm$0.09} & \dashuline{0.93$\pm$0.07} & 1.09$\pm$0.10 & 1.07$\pm$0.04 & 1.00$\pm$0.04 & 1.00$\pm$0.01 & \underline{0.73$\pm$0.02} & \underline{0.82$\pm$0.01} & 0.83$\pm$0.09 & 0.89$\pm$0.06 & \textbf{0.68$\pm$0.03}&\textbf{0.78$\pm$0.03}\\
\midrule
\multirow{4}{*}{\rotatebox{90}{ETTm2}} &96 &1.00$\pm$0.13 & 1.00$\pm$0.08 & 0.93$\pm$0.05 & 0.94$\pm$0.05 & 0.93$\pm$0.04 & 0.94$\pm$0.02 & {0.94$\pm$0.08} & {0.97$\pm$0.05} & 0.96$\pm$0.19 & 0.95$\pm$0.10 & \dashuline{0.81$\pm$0.08} & \dashuline{0.85$\pm$0.05} & 1.00$\pm$0.11 & 1.00$\pm$0.07 & \underline{0.72$\pm$0.07} & \underline{0.78$\pm$0.02} & {0.79$\pm$0.14} & {0.84$\pm$0.08} & \textbf{0.70$\pm$0.07}&\textbf{0.77$\pm$0.05} \\ 
& 192 &1.00$\pm$0.11 & 1.00$\pm$0.05 & 1.01$\pm$0.11 & 1.01$\pm$0.06 & 1.13$\pm$0.26 & 1.07$\pm$0.14 & 0.92$\pm$0.06 & 0.96$\pm$0.02 & \dashuline{0.70$\pm$0.05} & \dashuline{0.82$\pm$0.04} & 0.81$\pm$0.15 & 0.90$\pm$0.08 & 0.99$\pm$0.09 & 1.00$\pm$0.04 & \textbf{0.54$\pm$0.00} & \textbf{0.73$\pm$0.01} & 0.76$\pm$0.02 & 0.87$\pm$0.02 & \underline{0.55$\pm$0.02}&\underline{0.74$\pm$0.02} \\
& 336 &1.00$\pm$0.05 & 1.00$\pm$0.02 & 0.94$\pm$0.05 & 0.98$\pm$0.02 & 0.92$\pm$0.12& 0.96$\pm$0.08 & 0.93$\pm$0.04 & 0.96$\pm$0.03 & \dashuline{0.85$\pm$0.02} &\dashuline{0.93$\pm$0.02} & 0.96$\pm$0.08 & 0.98$\pm$0.05 & 1.04$\pm$0.07 & 1.02$\pm$0.03 & \underline{0.62$\pm$0.13}&\underline{0.78$\pm$0.07} &{0.85$\pm$0.07} &{0.93$\pm$0.03} &\textbf{0.54$\pm$0.02}&\textbf{0.73$\pm$0.02} \\
& 720 & 1.00$\pm$0.08 & 1.00$\pm$0.05 & 1.26$\pm$0.08 & 1.15$\pm$0.06 & 1.38$\pm$0.01 &1.16$\pm$0.00& 0.95$\pm$0.09 & 0.97$\pm$0.06 & 0.94$\pm$0.10 &\dashuline{0.95$\pm$0.05} &\dashuline{0.90$\pm$0.21} &0.95$\pm$0.12 & 1.00$\pm$0.08 & 1.00$\pm$0.06 &\textbf{0.76$\pm$0.14} &\textbf{0.84$\pm$0.07} & 1.09$\pm$0.11 & 1.06$\pm$0.07 & \underline{0.81$\pm$0.07}&\underline{0.89$\pm$0.03}\\
\midrule
\multirow{4}{*}{\rotatebox{90}{Exchange}} &96 &1.00$\pm$0.04 & 1.00$\pm$0.02 & \dashuline{0.83$\pm$0.06} &\dashuline{0.88$\pm$0.03} &1.00$\pm$0.04 & 1.01$\pm$0.02 & 0.90$\pm$0.05 & 0.95$\pm$0.02 & 0.96$\pm$0.20 & 0.97$\pm$0.10 & 0.98$\pm$0.07 & 0.99$\pm$0.04 & 1.00$\pm$0.04 &1.00$\pm$0.02 & \underline{0.30$\pm$0.01} &\underline{0.51$\pm$0.00} & 0.84$\pm$0.02 & 0.90$\pm$0.01 & \textbf{0.28$\pm$0.00} &\textbf{0.50$\pm$0.01} \\ 
& 192 &1.00$\pm$0.01 & 1.00$\pm$0.00 & \dashuline{0.97$\pm$0.00} & \dashuline{0.94$\pm$0.00} & 1.06$\pm$0.01 & 1.04$\pm$0.01 & 1.03$\pm$0.04 & 1.02$\pm$0.02 & 1.06$\pm$0.02 & 1.02$\pm$0.01 & 1.00$\pm$0.02 & 1.02$\pm$0.01 & 1.00$\pm$0.01 & 1.00$\pm$0.00 &\underline{0.56$\pm$0.04} &\underline{0.74$\pm$0.03} & 1.01$\pm$0.04 & 0.98$\pm$0.02& \textbf{0.55$\pm$0.03} &\textbf{0.72$\pm$0.03} \\
& 336 &1.00$\pm$0.02 & 1.00$\pm$0.02 & 1.06$\pm$0.04 & 0.99$\pm$0.02 & 0.97$\pm$0.02 & 1.00$\pm$0.00 & 0.96$\pm$0.01 & 0.98$\pm$0.01 & 1.00$\pm$0.06 & 0.99$\pm$0.04 & \dashuline{0.91$\pm$0.07} & \dashuline{0.96$\pm$0.03} & 0.99$\pm$0.02 & 1.00$\pm$0.02 &\underline{0.63$\pm$0.08} &\underline{0.80$\pm$0.05} &0.90$\pm$0.01 & 0.94$\pm$0.01 & \textbf{0.58$\pm$0.08}&\textbf{0.77$\pm$0.05} \\
& 720 & 1.00$\pm$0.06 & 1.00$\pm$0.03 & 1.12$\pm$0.03  & 1.07$\pm$0.01 & 0.99$\pm$0.07 & 0.99$\pm$0.05 & \dashuline{0.87$\pm$0.08} & \dashuline{0.93$\pm$0.04} & 1.02$\pm$0.02 & 1.02$\pm$0.00 & 0.90$\pm$0.06 & 0.94$\pm$0.05 & 1.00$\pm$0.06 & 1.00$\pm$0.03 &\textbf{0.50$\pm$0.08} & \textbf{0.67$\pm$0.06} & {0.72$\pm$0.06} & 0.83$\pm$0.04 & \underline{0.62$\pm$0.05}&\underline{0.76$\pm$0.05} \\
\bottomrule
\end{tabular}
\label{tab:main-std-scale}
}
\end{footnotesize}
\vspace{-1.5em}
\end{table*}
\vspace{-10mm}
\end{document}